# Low-Resource Clickbait Spoiling for Indonesian via Question Answering

Ni Putu Intan Maharani[1], Ayu Purwarianti[1], Alham Fikri Aji[2]
[1]School of Electrical Engineering and Informatics
Institut Teknologi Bandung
Bandung, Indonesia
23522048@std.stei.itb.ac.id, ayu@itb.ac.id
[2]Natural Language Processing Department
Mohamed bin Zayed University of Artificial Intelligence
Abu Dhabi, UAE
alham.fikri@mbzuai.ac.ae

*Abstract*—Clickbait spoiling aims to generate a short text to satisfy the curiosity induced by a clickbait post. As it is a newly introduced task, the dataset is only available in English so far. Our contributions include the construction of manually labeled clickbait spoiling corpus in Indonesian and an evaluation on using cross-lingual zero-shot question answering-based models to tackle clikcbait spoiling for low-resource language like Indonesian. We utilize selection of multilingual language models. The experimental results suggest that XLM-RoBERTa (large) model outperforms other models for phrase and passage spoilers, meanwhile, mDeBERTa (base) model outperforms other models for multipart spoilers.

## I. INTRODUCTION

Clickbait spoiling in English is newly introduced by Hagen et al. [1] with the aim to generate a short text to satisfy the curiosity induced by a clickbait post. Three types of spoilers, namely: phrase, passage, and multipart are identified. They tackled the task, mainly for phrase and passage type, by using reading comprehension techniques (i.e., question answering and passage retrieval). For both spoiler types, they found that question answering model DeBERTa-large that was fine-tuned on SQuAD v1.1 [2] and Webis Clickbait Spoiling Corpus 2022 [1] outperformed all other models.

Indonesian is the fourth most widely utilized language on the internet, with approximately 212 million users worldwide[1]. It is also worth noting that Indonesia has a lot of online news platforms with increasing number of readers over the years. As they rely on number of clicks and visits from online readers, they might use clickbait titles or headlines to attract more readers. Moreover, according to study conducted by Connecticut State University in 2016[2], Indonesia is ranked 60th out of 61st for the world's most literate nations, meaning that Indonesia has a low reading interest rate. With the increasing number of clickbait posts and low reading interest rate, readers can be more susceptible to believing and sharing clickbait content without verifying its accuracy.

As clickbait spoiling data is only available in English up to this point, implementing clickbait spoiling task in Indonesian can be challenging. In this paper, we aim to extend the task of clickbait spoiling, specifically, its implementation for Indonesian as a low-resource language. This paper reports about our investigation into clickbait spoiling for low-resource language and the following contributions: (1) Indonesian Clickbait Spoiling Corpus, consisting of 629 clickbait titles, their linked pages, and a spoiling piece of text therein. (2) A systematic evaluation of question answering model to generate spoilers given the low-resource scenario.

## II. RELATED WORK

In this section, we present an overview of research on clickbait spoiling, question answering task, and methods of tackling question answering task in low-resource settings.

### A. Clickbait Spoiling

Clickbait spoiling task is introduced by Hagen et al. [1] in which it is implemented in English. They also constructed a dataset for clickbait spoiling called Webis Clickbait Spoiling Corpus 2022 that has 3200 training data, 800 validation data, and 1000 test data. As mentioned, Hagen et al. [1] used reading comprehension approaches namely question answering and passage retrieval to generate a spoiler from a clickbait post in an extractive manner. For the question answering models, they used ALBERT [3], AllenAIDocument-QA [4], BERT (cased/uncased) [5], Big Bird [6], DeBERTa (large) [7], ELECTRA [8], FunnelTransformer [9], MPNet [10], and RoBERTa (base/large) [11]. All of those language models were fine tuned by using SQuAD v1.1 [2] and further fine-tuned by using Webis Clickbait Spoiling Corpus 2022 [1]. For the passage retrieval models, they used MonoBERT [12] [13] and MonoT5 [14], BM25 [15] and Query Likelihood [16]. All of those models were trained by using MS MARCO dataset [17] and further fine-tuned by using Webis Clickbait Spoiling Corpus 2022 [1].

---
[1]https://www.internetworldstats.com/stats3.htm
[2]https://www.thejakartapost.com/life/2016/08/29/indonesia-ranks-second-last-in-reading-interest-study.html

For quantitative evaluation metrics, Hagen et al. [1] used BLEU-4 [18], METEOR [19] in its extended version of [20], BERTScore [21], and Precision@1. It was found that DeBERTa-large question answering model outperformed all other models for both phrase and passage spoilers.

In our experiments, we use multilingual language models, namely, multilingual BERT [5], XLM RoBERTa (both base and large architectures) [22], and multilingual DeBERTa [7]. Those models are first fine-tuned for question answering task by using SQuAD v2.0 [23] because of its improved dataset quality over the previous version.

### B. Question Answering

Question answering involves providing answers to questions, usually those related to reading comprehension. Question answering models are implemented by fine-tuning pre-trained language models using question answering dataset like SQuAD [2] [23]. Rajpurkar et al [2] compile more than 100,000 questions and answers based on 536 Wikipedia articles. In those 107,785 collected entries, 93.6% of the answers are factual (8.9% date, 10.9% other numeric, 12.9% person, 4.4% location, 15.3% other entity, 31.8% common noun phrase, 3.9% adjective phrase, and 5.5% verb phrase), while the remainder are descriptive (3.7% clauses and 2.7% other). In addition to the 107,785 entries in SQuAD, Rajpurkar et al [23], in SQuAD v2.0, add more 50,000 unanswerable questions if only based on the given context while still referencing entities in the paragraphs. We train the base of our clickbait spoiling models using SQuAD v2.0 because of its wide range and its newly added feature to answer questions that are not expressed explicitly in the given context. We argue that this feature will provide significant contribution in our clickbait spoiler models as the dataset itself may contain questions (clickbait posts or titles) that have clickbait characteristics (e.g., exaggeration, ambiguous or vague language) and may not stated directly in the associated content.

### C. Low-Resource Question Answering

Question answering task, and other downstream tasks, for low-resource languages, can be implemented by performing fine-tuning on multilingual contextual pretrained language model. The training data can be from high-resource language such as English and then the model is tested by using test data from low-resource language (cross-lingual zero-shot approach) [24] [25] [26] [27] [28]. Moreover, adding a small set of labeled data from low-resource language for training can also boost the performance of the model (few-shot learning) [29] [30] [31].

To tackle clickbait spoiling for Indonesian as a low-resource language, we experiment on performing cross-lingual zero-shot learning. For the zero-shot learning scenario, multilingual language models that are fine-tuned with SQuAD v2.0 [23] are further fine-tuned by using Webis Clickbait Spoiling Corpus 2022 [1]. Models resulted from scenario above are tested by using our Indonesian clickbait spoiling corpus.

## III. INDONESIAN CLICKBAIT SPOILING CORPUS

### A. Corpus Construction

Our corpus is primarily based on IndoSUM test data that consists of online news articles in Indonesian [32]. As it does not have the title for each news article, web scraping by using BeautifulSoup tool in Python is performed to retrieve the titles for the corresponding source urls. Then, we filter the entries to include only clickbait titles. We do so by inferencing those titles to a trained clickbait binary classifier. For the classifier, we use IndoBERT [33] and Indonesian clickbait classification dataset, CLICK-ID, by William and Sari [34]. The classifer model was trained for 3 epochs with learning rate of 1e-6 and dropout rate of 0.3. From 18,774 entries in IndoSUM test set, it is filtered to 2,081 clickbait entries. Figure 1 shows the amount of clickbait entries in IndoSUM test set grouped by several news platform sources, namely, CNN Indonesia, Merdeka, Suara, Juara.net, Dailysocial id, Goal Indonesia, Antara News, Kumparan, Poskota News, and Rima News. Entries from Kumparan, Poskota News, and Rima News are filtered out completely by our clickbait classifier.

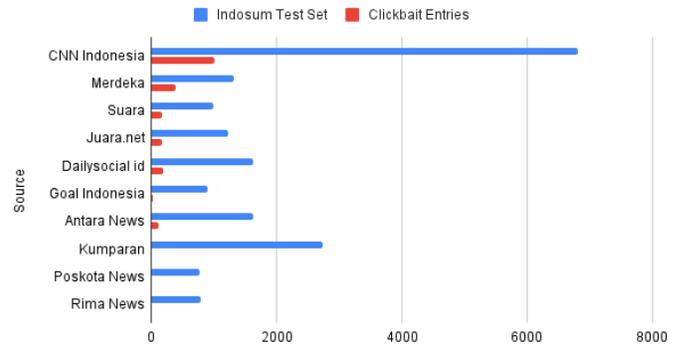

Fig. 1. Clickbait Entries in IndoSUM's Test Set by Source

With the target of 500 entries, we then annotate the spoilers for each entry in an extractive manner and identify the type of the extracted spoilers, namely, phrase, passage, or multipart. We also labeled their exact positions in the linked documents. As mentioned in previous research regarding clickbait spoiling corpus construction, a spoiler should be as short as possible (i.e., if one word is enough, not a whole sentence should be chosen) [1]. Since the spoiler annotation task is simple, one main annotator was sufficient. However, the annotated entries were reviewed and validated by two other validators. When both validators disagree with the extracted spoilers, those entries are dropped. Meanwhile, when at least one validator agrees with the extracted spoilers, those entries are re-annotated until both validators agree. We first annotate and validate 521 entries. Then, after applying the validation scenario mentioned, we drop 22 entries as both validators disagree with the annotated spoilers. The primary reason for the disagreement is that the spoilers did not quite fulfil what the clickbait titles "promise" in their contents. We also re-annotate 8 entries as only one validator agrees with the annotated spoilers. To reach our target, we annotate more entries with the total of 635 entries and after applying the validation scenario once again, we are left with 629 annotated and validated entries, more than 500 entries as targeted.

In sum, each of the 629 entries in our corpus consists of a unique ID, the online news platform which it was taken, the category of the news, the clickbait title, the source url, the main content of the linked document or news, the manually extracted spoiler including its position in the main content, and the type of spoiler (phrase, passage, or multipart)[3]. In total, the annotation took about 50 hours.

*B. Corpus Statistics*

Table I summarizes the main statistics of our corpus including the average text length of each dataset component (number of characters). Most of the spoiled clickbait titles come from CNN Indonesia (44%), followed by Merdeka (23%) and Suara (11%). Most spoilers are passages (51%) and phrases (34%). We do not provide a train/test split in this case as we only utilize the constructed data as a test data.

An essential aspect to consider is the vocabulary count of the constructed dataset. It serves as a fundamental indicator of the dataset's linguistic diversity and complexity. A larger vocabulary count often implies a richer and more varied language representation within the dataset. As shown in table II, out of 629 entries, the vocabulary count for clickbait titles, contents, and spoilers are 2,561, 18,160, and 2,522 respectively.

## IV. CLICKBAIT SPOILING: SPOILER GENERATION

From the study conducted by Hagen et al. [1], question answering models outperform passage retrieval model for both phrase and passage spoilers. Therefore, in this study, we only experiment on utilizing question answering for generating spoilers.

We view a clickbait title for both phrase and passage spoiler type as a "question" and the linked document/content as potentially containing an "answer" or the spoiler. We therefore employ four question answering methods trained on the SQuAD v2.0 data [23] and fine-tune them using the training and validation set of Webis Clickbait Spoiling Corpus 2022 [1]. For the language models, we use the ones that were pre-trained using multilingual corpus so that it can be used for inferencing our constructed clickbait spoiling corpus in Indonesian language. The models used namely, mBERT (base), XLM-R (both base and large architectures), and mDeBERTa (base). The multilingual question answering models are further fine-tuned using Webis Clickbait Spoiling Corpus 2022 [1] for 10 epochs, using learning rate of 2e-5, batch size of 16, and weight decay of 0.01. The fine-tuning is performed using Nvidia A100 GPUs provided by Google Colab.

[3]https://huggingface.co/datasets/intanm/indonesian-clickbait-spoiling

TABLE I
INDONESIAN CLICKBAIT SPOILING CORPUS DATASET STATISTICS

| Source | Spoiler Type | Entries | Average Text Length | | |
|---|---|---|---|---|---|
| | | | Clickbait Title | Context | Spoiler |
| CNN Indonesia | Phrase | 89 | 52.52 | 2116.45 | 16.76 |
| | Passage | 132 | 50.24 | 2126.88 | 77.58 |
| | Multipart | 46 | 46.02 | 2655.04 | 126.89 |
| Merdeka | Phrase | 45 | 57.04 | 1492.73 | 15.84 |
| | Passage | 78 | 55.85 | 1635.24 | 76.05 |
| | Multipart | 24 | 56.67 | 1771.38 | 95.63 |
| Suara | Phrase | 20 | 52.35 | 1939.95 | 19.25 |
| | Passage | 39 | 50.38 | 1978.77 | 82.90 |
| | Multipart | 13 | 46.31 | 2160.54 | 126.62 |
| Juara.net | Phrase | 25 | 57.40 | 1677.68 | 17.72 |
| | Passage | 27 | 52.74 | 1529.78 | 80.63 |
| | Multipart | 4 | 53.75 | 1756.75 | 154.00 |
| Daily Social id | Phrase | 23 | 69.13 | 2385.17 | 15.17 |
| | Passage | 22 | 66.68 | 3003.68 | 100.55 |
| | Multipart | 4 | 67.25 | 2581.25 | 173.00 |
| Goal Indonesia | Phrase | 1 | 49.00 | 1980.00 | 15.00 |
| | Passage | 5 | 53.80 | 1194.40 | 54.00 |
| | Multipart | - | - | - | - |
| Antara News | Phrase | 13 | 45.23 | 1779.54 | 18.92 |
| | Passage | 16 | 49.25 | 1701.31 | 82.06 |
| | Multipart | 3 | 35.67 | 2070.33 | 93.00 |
| Total | Phrase | 216 | 55.32 | 1927.09 | 16.87 |
| | Passage | 319 | 52.98 | 1962.53 | 79.55 |
| | Multipart | 94 | 49.68 | 2301.01 | 120.90 |
| | All | 629 | 53.29 | 2000.94 | 64.21 |

TABLE II
DATASET VOCABULARY COUNT

| Dataset Component | Vocab Count |
|---|---|
| Clickbait Title | 2,561 |
| Content | 18,160 |
| Spoiler | 2,522 |

## V. EVALUATION OF SPOILER GENERATION

We assess the effectiveness of the question answering methods for clickbait spoiling on all of the spoiler types. We also include multipart spoiler in the evaluation even though there might be no exact spoilers detected, in addition, it is to show if there are partial matches. To assess the quantitative correspondence between a predicted spoiler and the ground truth, we use BLEU [18], METEOR [19] in its extended version [20], BERTScore [21], and SQuAD [2] measures.

For all measures, we use the library provided by Huggingface in which the all take reference spoilers and predicted spoilers as inputs. In the case of BERTScore, we set the `lang` parameter to `id` for Indonesian.

From Table III, it can be observed that fine-tuned XLM RoBERTa (large) outperforms all other models for all spoiler types. It has an exact match score of 22.417% which means the model manages to predict the spoilers of 141 entries (out of 629) exactly as the ground truth. Meanwhile, the predicted spoilers for approximately 261 entries are both exactly and partially matched to the ground truth.

*A. Effectiveness on Phrase Spoilers*

Table IV shows the effectiveness of the selected question answering models on 216 clickbait titles with phrase spoil-ers. Given the ground-truth spoiler, we report the predicted spoilers' SQuAD's Exact Match, F1, BLEU, METEOR, and BERTScore.

TABLE III
EXPERIMENT RESULTS FOR ALL SPOILER TYPES

| Model | SQuAD EM | SQuAD F1 | BERT Score | BLEU | METEOR |
|---|---|---|---|---|---|
| mBERT (base) | 15.421 | 30.713 | 74.015 | 8.266 | 24.836 |
| XLM RoBERTa (base) | 20.191 | 35.604 | 75.955 | 9.687 | 30.210 |
| mDeBERTa (base) | 18.760 | 35.591 | 75.763 | 8.053 | 29.345 |
| XLM RoBERTa (large) | **22.417** | **41.519** | **77.687** | **12.838** | **34.954** |

TABLE IV
EXPERIMENT RESULTS FOR PHRASE SPOILERS

| Model | SQuAD EM | SQuAD F1 | BERT Score | BLEU | METEOR |
|---|---|---|---|---|---|
| mBERT (base) | 37.963 | 49.391 | 82.295 | 22.465 | 43.984 |
| XLM RoBERTa (base) | 47.685 | 57.795 | 84.872 | 25.081 | 53.449 |
| mDeBERTa (base) | 45.833 | 57.905 | 84.845 | 25.262 | 52.352 |
| XLM RoBERTa (large) | **53.241** | **66.804** | **87.773** | **30.252** | **60.427** |

Overall, XLM RoBERTa (large) outperforms all other models for phrase spoilers. Based on the SQuAD's Exact Match score, it predicts the correct spoiler for 115 out of 216 test data (i.e., for about 53.241%). According to BERTScore, it achieves the similarity score of 87.773%, however, it has a score of 30.252% and 60.427% on BLEU and METEOR respectively. This indicates that the model excels in capturing semantic similarity (leverages contextual embeddings from BERTScore), while relatively low scores on METEOR and BLEU suggest that it may struggle with more specific n-gram matching and another matching criteria emphasized by these measures.

TABLE V
EXPERIMENT RESULTS FOR PASSAGE SPOILERS

| Model | SQuAD EM | SQuAD F1 | BERT Score | BLEU | METEOR |
|---|---|---|---|---|---|
| mBERT (base) | 4.702 | 17.925 | 69.153 | 6.461 | 14.579 |
| XLM RoBERTa (base) | 7.524 | 21.994 | 71.020 | 8.647 | 18.941 |
| mDeBERTa (base) | 5.956 | 21.479 | 70.570 | 6.142 | 17.497 |
| XLM RoBERTa (large) | **8.150** | **27.735** | **72.613** | **12.482** | **23.684** |

TABLE VI
EXPERIMENT RESULTS FOR MULTIPART SPOILERS

| Model | SQuAD EM | SQuAD F1 | BERT Score | BLEU | METEOR |
|---|---|---|---|---|---|
| mBERT (base) | 0.000 | 31.196 | 71.487 | 1.050 | 15.645 |
| XLM RoBERTa (base) | 0.000 | 30.800 | 72.214 | 0.585 | 15.050 |
| mDeBERTa (base) | 0.000 | **32.210** | **72.514** | 1.130 | **16.687** |
| XLM RoBERTa (large) | 0.000 | 30.192 | 71.729 | **1.276** | 14.667 |

### B. Effectiveness on Passage Spoilers

Table V shows the effectiveness of the selected question answering models on the 319 clickbait titles with passage spoilers. As we can see, the numbers are lower for all models compared to the phrase spoilers. While it shows low scores on SQuAD's Exact Match, F1, BLEU, and METEOR, most models managed to reach above 70% of contextual similarity. Overall, XLM RoBERTa (large) also outperforms all other models. In fact, it has quite a small margin with the XLM RoBERTa (base) model in terms of SQuAD's Exact Match score (0.627%).

### C. Effectiveness on Multipart Spoilers

As mentioned in Hagen et al. [1], we need to use other approach to tackle multipart spoilers due to its non-consecutive spoiler positions. However, even though we know there will be no exact matches for the predicted spoilers, the results shown in Table VI indicates that the models are able to capture partial matches (from SQuAD's F1) and relatively high contextual similarity (from BERTScore). Interestingly, mDeBERTa model outperforms all other models for multipart spoilers in three of the measures (i.e., SQuAD's F1, BERTScore, and METEOR).

### D. Source-based Evaluation

The dataset is gathered from multiple news platforms, namely, CNN Indonesia, Merdeka, Suara, Juara.net, Dailysocial id, Goal Indonesia, and Antara News. Detailed statistics of each source can also be seen in Table I. Due to its limited representation in the dataset, we exclude Goal Indonesia and Antara News for this evaluation (less than 40 entries).

As we can see in Table VII, for SQuAD Exact Match measure, the models can predict exact match spoilers for the entries gathered from Dailysocial id better than other sources with the average performance of 25.510%. For SQuAD F1 measure, predicted spoilers for Dailysocial id entries are also better than other sources. In terms of semantic similarity measured by BERTScore, the models predict semantically similar spoilers for entries from Suara better than others. For BLEU measure, spoilers generated for entries from Juara.net are better than other sources. In terms of METEOR metric, the predicted spoilers for Suara outperform other sources.

Question answering model works well with shorter answers to be predicted, that is why Hagen et al [1] used passage retrieval models to handle longer spoilers (i.e., passage spoilers) even though question answering models still outperformed the passage retrieval models. In addition, it is shown in this research that the question answering models predict phrase spoilers way better than other types with longer spoilers. In Table I, we can see that the number of each spoiler type varies throughout the sources in which most sources have more passage spoilers (longer text to be predicted). The number of phrase and passage spoilers for both Dailysocial id and Juara.net are close and thus it leads to a better performance for both sources (SQuAD metrics, BLEU, METEOR).

TABLE VII
EXPERIMENT RESULTS FOR ALL SPOILERS BY SOURCE

| Model | Metric | CNN Indonesia | Merdeka | Suara | Juara.net | Dailysocial id |
|---|---|---|---|---|---|---|
| mBERT (base) | SQuAD EM | 16.105 | 14.966 | 6.944 | 12.5 | 24.49 |
| XLM RoBERTa (base) | SQuAD EM | 16.854 | 22.449 | 13.888 | 26.786 | 24.490 |
| mDeBERTa (base) | SQuAD EM | 18.727 | 20.408 | 12.500 | 19.643 | 20.408 |
| XLM RoBERTa (large) | SQuAD EM | 22.097 | 25.170 | 12.500 | 25.170 | 32.653 |
| **AVERAGE** | | 18.446 | 20.748 | 11.458 | 21.025 | **25.510** |
| mBERT (base) | SQuAD F1 | 30.714 | 30.124 | 28.733 | 27.296 | 35.364 |
| XLM RoBERTa (base) | SQuAD F1 | 32.978 | 38.374 | 33.599 | 38.008 | 37.142 |
| mDeBERTa (base) | SQuAD F1 | 34.566 | 35.888 | 32.635 | 39.989 | 35.871 |
| XLM RoBERTa (large) | SQuAD F1 | 40.208 | 46.187 | 35.139 | 46.187 | 47.124 |
| **AVERAGE** | | 34.617 | 37.643 | 32.527 | 37.870 | **38.875** |
| mBERT (base) | BERT Score | 74.351 | 75.362 | 74.110 | 68.878 | 76.315 |
| XLM RoBERTa (base) | BERT Score | 75.999 | 77.382 | 78.046 | 73.986 | 75.999 |
| mDeBERTa (base) | BERT Score | 75.797 | 76.376 | 77.305 | 72.980 | 74.538 |
| XLM RoBERTa (large) | BERT Score | 77.764 | 79.216 | 81.532 | 74.885 | 74.501 |
| **AVERAGE** | | 75.978 | 77.084 | **77.749** | 72.682 | 75.338 |
| mBERT (base) | BLEU | 7.014 | 7.937 | 12.021 | 10.696 | 5.254 |
| XLM RoBERTa (base) | BLEU | 8.262 | 10.636 | 12.625 | 8.929 | 7.578 |
| mDeBERTa (base) | BLEU | 7.937 | 6.511 | 9.282 | 10.353 | 4.851 |
| XLM RoBERTa (large) | BLEU | 11.433 | 13.971 | 11.946 | 16.237 | 11.969 |
| **AVERAGE** | | 8.662 | 9.764 | 11.469 | **11.554** | 7.413 |
| mBERT (base) | METEOR | 25.889 | 29.307 | 22.692 | 10.570 | 27.554 |
| XLM RoBERTa (base) | METEOR | 31.098 | 33.220 | 36.006 | 23.270 | 21.133 |
| mDeBERTa (base) | METEOR | 30.710 | 32.304 | 30.569 | 17.749 | 21.335 |
| XLM RoBERTa (large) | METEOR | 35.116 | 37.028 | 46.784 | 24.059 | 25.572 |
| **AVERAGE** | | 30.703 | 32.965 | **34.013** | 18.912 | 23.898 |

## VI. CONCLUSION AND FUTURE WORKS

We compile the first resource for clickbait in Indonesian with associated spoilers. We interpret the clickbait spoiling task for all spoiler types as a question answering task and there are many possible approaches to tackle clickbait spoiling. We have performed cross-lingual zero-shot experiments on multilingual question answering models (trained and validated using Webis Clickbait Spoiling Corpus 2022 [1]) and evaluate their effectiveness for type-specific clickbait spoiling using our constructed clickbait spoiling corpus in Indonesian. We also evaluate the performance of our clickbait spoiling models for each news platform source of the corpus entries. Overall, our results show that question answering-based approach is effective in generating spoilers, especially for phrase spoiler type. Moreover, the utilization of multilingual pre-trained language models is also effective to tackle the task for low-resource language like Indonesian.

With respect to multipart spoilers, several approaches such as question answering for multi-span answers and summarization models could be interesting directions to extract non-consecutive spoilers from the linked content of a clickbait title. With also a focus on tackling NLP task for low-resource language, there might be various approaches other than cross-lingual zero-shot learning: for example, few-shot learning that incorporates a small scale of training data in target language, and semi-supervised learning that leverages unlabeled data for training.

## ACKNOWLEDGMENT

We sincerely appreciate Adi Susilayasa and Giri Dharma to help validate the constructed dataset. We also would like to thank Rinaldi Sirait and Khumaeni for sharing their insights about this project.

## REFERENCES

[1] Matthias Hagen, Maik Fröbe, Artur Jurk, and Martin Potthast. Clickbait spoiling via question answering and passage retrieval. In *Proceedings of the 60th Annual Meeting of the Association for Computational Linguistics (Volume 1: Long Papers)*, pages 7025–7036, Dublin, Ireland, May 2022. Association for Computational Linguistics.

[2] Pranav Rajpurkar, Jian Zhang, Konstantin Lopyrev, and Percy Liang. SQuAD: 100,000+ questions for machine comprehension of text. In *Proceedings of the 2016 Conference on Empirical Methods in Natural Language Processing*, pages 2383–2392, Austin, Texas, November 2016. Association for Computational Linguistics.

[3] Zhenzhong Lan, Mingda Chen, Sebastian Goodman, Kevin Gimpel, Piyush Sharma, and Radu Soricut. Albert: A lite bert for self-supervised learning of language representations, 2020.

[4] Christopher Clark and Matt Gardner. Simple and effective multi-paragraph reading comprehension. In *Proceedings of the 56th Annual Meeting of the Association for Computational Linguistics (Volume 1: Long Papers)*, pages 845–855, Melbourne, Australia, July 2018. Association for Computational Linguistics.

[5] Jacob Devlin, Ming-Wei Chang, Kenton Lee, and Kristina Toutanova. BERT: Pre-training of deep bidirectional transformers for language understanding. In *Proceedings of the 2019 Conference of the North American Chapter of the Association for Computational Linguistics: Human Language Technologies, Volume 1 (Long and Short Papers)*, pages 4171–4186, Minneapolis, Minnesota, June 2019. Association for Computational Linguistics.

[6] Manzil Zaheer, Guru Guruganesh, Avinava Dubey, Joshua Ainslie, Chris Alberti, Santiago Ontanon, Philip Pham, Anirudh Ravula, Qifan Wang, Li Yang, and Amr Ahmed. Big bird: Transformers for longer sequences, 2021.

[7] Pengcheng He, Xiaodong Liu, Jianfeng Gao, and Weizhu Chen. Deberta: Decoding-enhanced bert with disentangled attention, 2021.

[8] Kevin Clark, Minh-Thang Luong, Quoc V. Le, and Christopher D. Manning. Electra: Pre-training text encoders as discriminators rather than generators, 2020.

[9] Zihang Dai, Guokun Lai, Yiming Yang, and Quoc V. Le. Funnel-transformer: Filtering out sequential redundancy for efficient language processing, 2020.

[10] Kaitao Song, Xu Tan, Tao Qin, Jianfeng Lu, and Tie-Yan Liu. Mpnet: Masked and permuted pre-training for language understanding, 2020.

[11] Yinhan Liu, Myle Ott, Naman Goyal, Jingfei Du, Mandar Joshi, Danqi Chen, Omer Levy, Mike Lewis, Luke Zettlemoyer, and Veselin Stoyanov. Roberta: A robustly optimized bert pretraining approach, 2019.

[12] Rodrigo Nogueira and Kyunghyun Cho. Passage re-ranking with bert, 2020.

[13] Rodrigo Nogueira, Wei Yang, Kyunghyun Cho, and Jimmy Lin. Multi-stage document ranking with bert, 2019.

[14] Rodrigo Nogueira, Zhiying Jiang, and Jimmy Lin. Document ranking with a pretrained sequence-to-sequence model, 2020.

[15] Stephen Robertson and Hugo Zaragoza. The probabilistic relevance framework: Bm25 and beyond. *Foundations and Trends® in Information Retrieval*, 3(4):333–389, 2009.

[16] Jay M. Ponte and W. Bruce Croft. A language modeling approach to information retrieval. *SIGIR Forum*, 51(2):202–208, aug 2017.

[17] Payal Bajaj, Daniel Campos, Nick Craswell, Li Deng, Jianfeng Gao, Xiaodong Liu, Rangan Majumder, Andrew McNamara, Bhaskar Mitra, Tri Nguyen, Mir Rosenberg, Xia Song, Alina Stoica, Saurabh Tiwary, and Tong Wang. Ms marco: A human generated machine reading comprehension dataset, 2018.

[18] Kishore Papineni, Salim Roukos, Todd Ward, and Wei-Jing Zhu. Bleu: a method for automatic evaluation of machine translation. In *Proceedings of the 40th Annual Meeting of the Association for Computational Linguistics*, pages 311–318, Philadelphia, Pennsylvania, USA, July 2002. Association for Computational Linguistics.

[19] Satanjeev Banerjee and Alon Lavie. METEOR: An automatic metric for MT evaluation with improved correlation with human judgments. In *Proceedings of the ACL Workshop on Intrinsic and Extrinsic Evaluation Measures for Machine Translation and/or Summarization*, pages 65–72, Ann Arbor, Michigan, June 2005. Association for Computational Linguistics.

[20] Michael Denkowski and Alon Lavie. Meteor universal: Language specific translation evaluation for any target language. In *Proceedings of the Ninth Workshop on Statistical Machine Translation*, pages 376–380, Baltimore, Maryland, USA, June 2014. Association for Computational Linguistics.

[21] Tianyi Zhang, Varsha Kishore, Felix Wu, Kilian Q. Weinberger, and Yoav Artzi. Bertscore: Evaluating text generation with bert, 2020.

[22] Alexis Conneau, Kartikay Khandelwal, Naman Goyal, Vishrav Chaudhary, Guillaume Wenzek, Francisco Guzmán, Edouard Grave, Myle Ott, Luke Zettlemoyer, and Veselin Stoyanov. Unsupervised cross-lingual representation learning at scale. In *Proceedings of the 58th Annual Meeting of the Association for Computational Linguistics*, pages 8440–8451, Online, July 2020. Association for Computational Linguistics.

[23] Pranav Rajpurkar, Robin Jia, and Percy Liang. Know what you don't know: Unanswerable questions for SQuAD. In *Proceedings of the 56th Annual Meeting of the Association for Computational Linguistics (Volume 2: Short Papers)*, pages 784–789, Melbourne, Australia, July 2018. Association for Computational Linguistics.

[24] Yu-Hsiang Lin, Chian-Yu Chen, Jean Lee, Zirui Li, Yuyan Zhang, Mengzhou Xia, Shruti Rijhwani, Junxian He, Zhisong Zhang, Xuezhe Ma, Antonios Anastasopoulos, Patrick Littell, and Graham Neubig. Choosing transfer languages for cross-lingual learning. In *Proceedings of the 57th Annual Meeting of the Association for Computational Linguistics*, pages 3125–3135, Florence, Italy, July 2019. Association for Computational Linguistics.

[25] Rasmus Hvingelby, Amalie Brogaard Pauli, Maria Barrett, Christina Rosted, Lasse Malm Lidegaard, and Anders Søgaard. DaNE: A named entity resource for Danish. In *Proceedings of the Twelfth Language Resources and Evaluation Conference*, pages 4597–4604, Marseille, France, May 2020. European Language Resources Association.

[26] Tsung-Yuan Hsu, Chi-Liang Liu, and Hung-yi Lee. Zero-shot reading comprehension by cross-lingual transfer learning with multi-lingual language representation model. In *Proceedings of the 2019 Conference on Empirical Methods in Natural Language Processing and the 9th International Joint Conference on Natural Language Processing (EMNLP-IJCNLP)*, pages 5933–5940, Hong Kong, China, November 2019. Association for Computational Linguistics.

[27] Lukas Lange, Anastasiia Iurshina, Heike Adel, and Jannik Strötgen. Adversarial alignment of multilingual models for extracting temporal expressions from text. In *Proceedings of the 5th Workshop on Representation Learning for NLP*, pages 103–109, Online, July 2020. Association for Computational Linguistics.

[28] Benjamin Muller, Benoit Sagot, and Djamé Seddah. Can multilingual language models transfer to an unseen dialect? a case study on north african arabizi, 2020.

[29] Anne Lauscher, Vinit Ravishankar, Ivan Vulić, and Goran Glavaš. From zero to hero: On the limitations of zero-shot language transfer with multilingual Transformers. In *Proceedings of the 2020 Conference on Empirical Methods in Natural Language Processing (EMNLP)*, pages 4483–4499, Online, November 2020. Association for Computational Linguistics.

[30] Michael A. Hedderich, David Adelani, Dawei Zhu, Jesujoba Alabi, Udia Markus, and Dietrich Klakow. Transfer learning and distant supervision for multilingual transformer models: A study on African languages. In *Proceedings of the 2020 Conference on Empirical Methods in Natural Language Processing (EMNLP)*, pages 2580–2591, Online, November 2020. Association for Computational Linguistics.

[31] Rakesh Chada and Pradeep Natarajan. FewshotQA: A simple framework for few-shot learning of question answering tasks using pre-trained text-to-text models. In *Proceedings of the 2021 Conference on Empirical Methods in Natural Language Processing*, pages 6081–6090, Online and Punta Cana, Dominican Republic, November 2021. Association for Computational Linguistics.

[32] Kemal Kurniawan and Samuel Louvan. Indosum: A new benchmark dataset for indonesian text summarization, 2019.

[33] Bryan Wilie, Karissa Vincentio, Genta Indra Winata, Samuel Cahyawijaya, Xiaohong Li, Zhi Yuan Lim, Sidik Soleman, Rahmad Mahendra, Pascale Fung, Syafri Bahar, and Ayu Purwarianti. IndoNLU: Benchmark and resources for evaluating Indonesian natural language understanding. In *Proceedings of the 1st Conference of the Asia-Pacific Chapter of the Association for Computational Linguistics and the 10th International Joint Conference on Natural Language Processing*, pages 843–857, Suzhou, China, December 2020. Association for Computational Linguistics.

[34] Andika William and Yunita Sari. Click-id: A novel dataset for indonesian clickbait headlines. *Data in Brief*, 32:106231, 2020.